\documentclass[conference]{IEEEtran}

\usepackage{amssymb,amsmath}
\usepackage{cite}
\usepackage{graphicx}
\usepackage{subfig}

\usepackage{psfrag}
\usepackage{url}
\usepackage[latin1]{inputenc}
\usepackage[absolute,overlay]{textpos}
\usepackage{gensymb}
\usepackage{cases}
\usepackage[font=footnotesize]{caption}
\usepackage{float}
\usepackage[linesnumbered,ruled,lined]{algorithm2e}
\newcommand{\argmax}[1]{{\operatorname{arg}\,\max_{#1}}\,}

\usepackage{pgf}

\usepackage[nocomma]{optidef}

\usepackage{amsthm}

\usepackage{booktabs}
\usepackage{color,soul}

\usepackage{textcomp}
\usepackage{bbm}
\usepackage{xcolor}
\def\BibTeX{{\rm B\kern-.05em{\sc i\kern-.025em b}\kern-.08em
    T\kern-.1667em\lower.7ex\hbox{E}\kern-.125emX}}

\usepackage{tabularx}
\usepackage{makecell}
\usepackage{multirow}

\begin{document}

\title{Offline and Distributional Reinforcement Learning for Radio Resource Management}
\author{
	\IEEEauthorblockN{Eslam Eldeeb and Hirley Alves \\
	}
	\IEEEauthorblockA{Centre for Wireless Communications (CWC), University of Oulu, Finland \\
	Email: firstname.lastname@oulu.fi}
}
\maketitle

\begin{abstract}
Reinforcement learning (RL) has proved to have a promising role in future intelligent wireless networks. Online RL has been adopted for radio resource management (RRM), taking over traditional schemes. However, due to its reliance on online interaction with the environment, its role becomes limited in practical, real-world problems where online interaction is not feasible. In addition, traditional RL stands short in front of the uncertainties and risks in real-world stochastic environments. In this manner, we propose an offline and distributional RL scheme for the RRM problem, enabling offline training using a static dataset without any interaction with the environment and considering the sources of uncertainties using the distributions of the return. Simulation results demonstrate that the proposed scheme outperforms conventional resource management models. In addition, it is the only scheme that surpasses online RL with a $10 \%$ gain over online RL.

\end{abstract}

\section{Introduction}\label{sec:introduction}

Recent advances towards $6$G networks include complex real-world wireless environments requiring continuous monitoring and control. Such advancements urge the need for new learning-based methods for modeling and control. Recently, reinforcement learning (RL) has become attractive to the wireless domain due to its outstanding ability to provide model-free decision-making~\cite{8714026}. In online RL, an agent observes the current state of the environment, takes a decision (action), transits to a new state, receives a feedback reward evaluating the decision, and improves its policy\footnote{A policy is a behavior that describes which actions are selected at each state.}. To this end, deploying RL to radio resource management (RRM) problems is appealing due to their complex optimization objectives and the settled monitoring of their environments through feedback signals~\cite{10004901}.

Several works have exploited RL techniques in the RRM problem~\cite{8761431,9679392,naderializadeh2021resource,10066838,8931561}. For instance, the work in~\cite{8761431} proposes a power allocation and resource management scheme using deep RL, achieving a high average sum rate in different user densities. The authors in~\cite{9679392} leverage RL to perform a distributed resource scheduling while minimizing the on-grid energy consumption. In~\cite{naderializadeh2021resource}, a multi-agent RL (MARL) solution to the RRM is presented, whereas a graph neural network (GNN) architecture is proposed to solve the RRM problem in~\cite{10066838}. In~\cite{8931561}, the authors propose a resource management algorithm in network slicing using a combination of generative adversarial networks (GANS) and deep RL. All in all, the majority of the literature, if not all, focused on online RL while solving the RRM problem.

Although online RL contributes greatly to solving RRM, it faces serious challenges when transferred to real-world problems. Online RL heavily relies on exploring the environment, which is a random exploration at the beginning of the optimization, through online interaction with the environments. Traditional online RL faces four main obstacles:
\begin{enumerate}
    \item random exploration at the beginning of the optimization introduces poor service to the users;

    \item random exploration wastes time and resources. In addition, it might not be safe to interact randomly with the environment;

    \item in complex applications, such as RRM, online RL needs a huge amount of interaction and takes long periods to converge, which might not be practically feasible; and

    \item traditional RL only considers the average performance, neglecting sources of uncertainties and risks.
\end{enumerate}
These challenges motivate the importance of offline RL and distributional RL. The former suggests optimizing the optimum policy using a static dataset without any interaction with the environment~\cite{levine2020offline}. This ensures safe optimization and cost minimization. In contrast, the latter considers the environment uncertainties relying on the return distribution instead of the average return~\cite{bdr2023}. This ensures mitigation sources of risks by considering the worst-case scenarios (tail of the return distribution).

This work proposes a novel offline and distributional RL algorithm for the RRM problem. In contrast, to~\cite{10529190}, which applies the offline RL scheme to the RRM, comparing a mixture of datasets, we rely on combining both offline and distributional RL to overcome the stochastic behavior of the environment. To our knowledge, this is the first work to combine offline RL and distributional RL for the RRM problem. The contributions of this paper are summarized as follows:
\begin{itemize}
    \item We propose an offline and distributional RL solution for the RRM problem. The proposed model maximizes the combination of the weighted sum and tail rates.

    \item We demonstrate the theoretical details of the proposed algorithm as well as practical implementation and dataset collection.
    
    \item We compared the proposed model to multiple benchmark schemes and an online RL algorithm.
        
    \item Numerical results show that the proposed offline and distributional RL scheme outperforms the state-of-the-art schemes. In addition, it is the only algorithm that surpasses online RL.
\end{itemize}

The rest of the paper is summarized as follows: Section~\ref{sec:system_model} presents the system model. Section~\ref{sec:RL} presents the RL formulation. The proposed offline and distributional RL algorithm is presented in Section~\ref{sec:ODRL}. Section~\ref{sec:results} depicts the results, and finally, Section~\ref{sec:conclusions} concludes the paper.

\vspace{-1mm}
\section{System Model}\label{sec:system_model}

\begin{figure}[t!]
    \centering    \includegraphics[width=1\columnwidth,trim={1cm 0 1cm 0},clip]{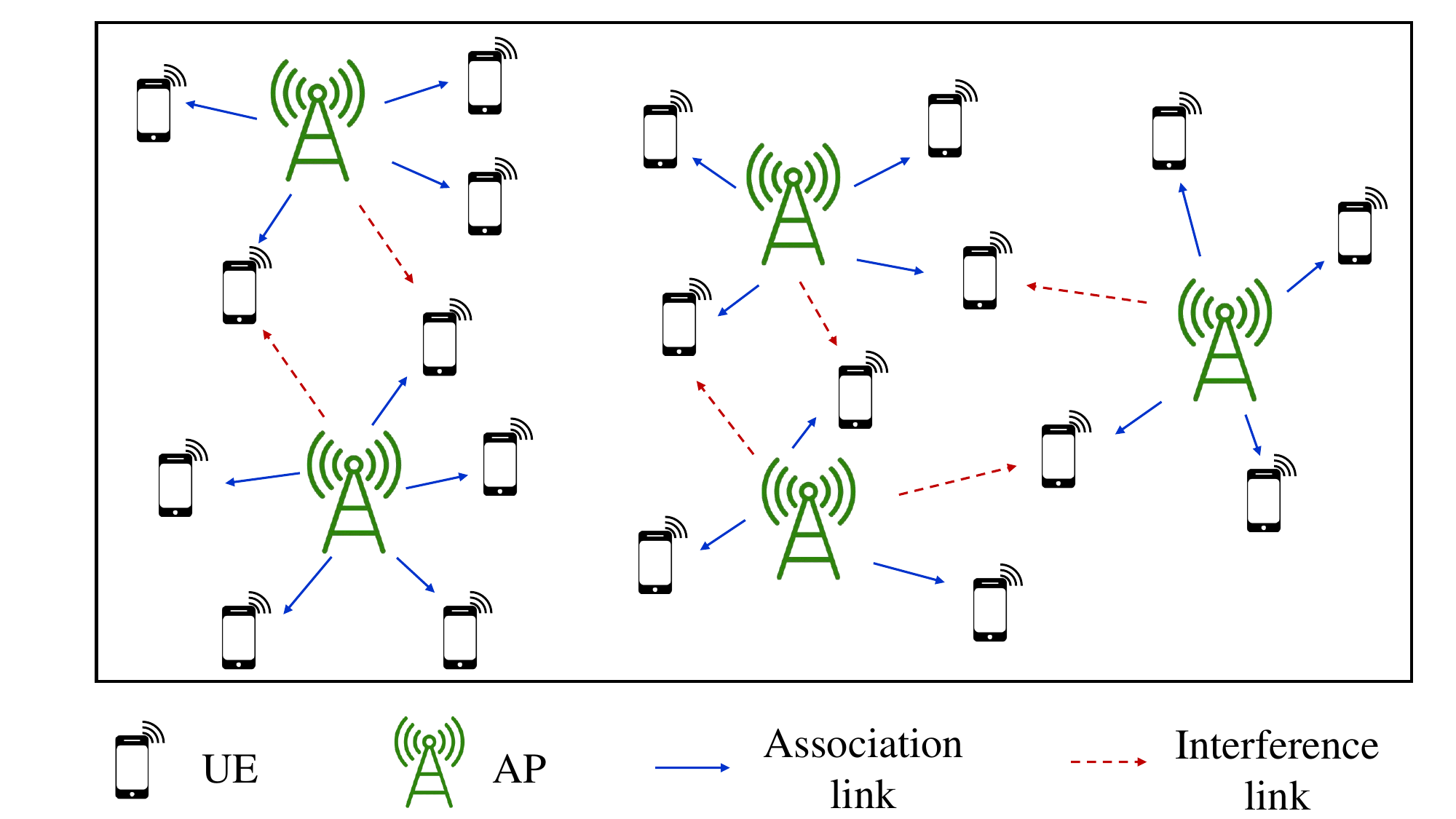} \vspace{-1mm}
    \caption{The wireless model consists of $N$ APs serving $M$ UEs. The blue lines represent the user association performed at the beginning of each episode, while the red lines represent interference links.}
    \vspace{-1mm}
    \label{sys_mod}
\end{figure}

Consider a wireless network of area $L \times L$ square meters that comprises $N$ randomly deployed access points (APs) and $M$ randomly deployed user equipment (UEs) as shown in Fig.~\ref{sys_mod}. To ensure practical scenarios, the random deployment of APs and UEs is controlled by minimum AP-UE distance $d_0$ and minimum AP-AP distance $d_1$. Our simulator considers an episodic environment consisting of time slots $t \in \{1, 2, \cdots, T\}$, where $T$ is the total number of slots in one episode. At each time slot $t$, each UE moves randomly in the network with a speed $v(t)$. The APs and UEs locations are fixed during each episode. 

At the beginning of each episode, user association occurs, where each UE is associated with one of the APs. User-association is performed according to the reference signal received power (RSRP) of each user, where UE $m$ is associated to AP $n$, such that $n = \argmax{i} \text{RSRP}_{mi}$, where $i \in \{1, 2, \cdots, N\}$. The RSRP is affected by the transmit power $p_t$ and the channel. The channel between UE $m$ and AP $n$, $h_{mn}$, is characterized by indoor path loss, log-normal shadowing with standard deviation $\sigma_{sh}$, and short-term frequency-flat Rayleigh fading. According to 3GPP~\cite{3gpp_indoor}, the indoor path loss between  UE $m$ and AP $n$ is calculated as
\begin{equation}
    \label{Indoor_PL}
    PL_{mn} = 15.3 + 37.6 \log(d_{mn}) + PL_o,
\end{equation}
where $d_{mn}$ is the euclidean distance between UE $m$ and AP $n$, $d_{mn} > d_0$ and  $PL_o$ is a reference path loss. The total power loss is the addition of the path loss, shadowing, and Rayleigh fading.

At time $t$, the received signal of UE $m$ that is associated with AP $n$ is
\begin{equation}
    \label{received_signal}
    y_m(t) = h_{mn}(t) \: x_n(t) + \sum_{i \neq n} h_{mi}(t) \: x_i(t) + n_m(t),
\end{equation}
where $n_m(t) \sim \mathcal{C} \mathcal{N} (0,n_o^2) $ is the additive white Gaussian noise (AWGN) with variance $n_o^2$. At time $t$, the instantaneous rate (Shannon capacity) of UE $m$ that is associated with AP $n$ is
\begin{equation}
    \label{inst_rate}
    R_m (t) = \log_2 \bigg(1 + \frac{|h_{mn}(t)|^2 \: p_t}{ \sum_{i \neq n} |h_{mi}(t)|^2 \: p_t + n_o^2}\bigg),
\end{equation}
where the term $\frac{|h_{mn}(t)|^2 \: p_t}{ \sum_{i \neq n} |h_{mi}(t)|^2 \: p_t + n_o^2}$ represents the signal-to-interference-plus-noise ratio (SINR) of UE $m$ at time $t$. The average rate (throughput) of UE $m$ in an episode is
\begin{equation}
    \Bar{R}_m = \frac{1}{T} \sum_{t=1}^T R_m (t).
\end{equation}

RRM problems aim to maximize the average data rates across all users. However, this objective is trivial to be solved in a way that each AP always schedules the user with the best SINR. Therefore, fairness across users must be considered in the RRM objective. In this work, we consider joint optimization of the sum-rate and the $5$-percentile rate. The former is calculated as
\begin{equation}
    \label{sum_rate}
    R_{\text{sum}} = \sum_{m = 1}^M \Bar{R}_m,
\end{equation}
where the latter is the average rate achieved by $95 \%$ of the UEs. The $5$-percentile rate is calculated as
\begin{align}
    \label{5_perc_rate}
    R_{5 \%} = \: &\max \: \: R \\
    & \: \: \: \text{s.t.} \: \: \: \: \mathbb{P} [\Bar{R}_m \geq R] \geq 0.95, \quad \forall m \in {1, 2, \cdots, M}. \nonumber
\end{align}
To this end, the objective is a weighted sum (joint combination) of the sum rate and the $5$-percentile rate
\begin{equation}
    \label{r_score}
    R_{\text{score}} = \mu_1 R_{\text{sum}} + \mu_2 R_{5 \%},
\end{equation}
where $\mu_1$ and $\mu_2$ are user-chosen weights.

\subsection{Problem Definition}
The objective in this work is user scheduling, \emph{i.e.}, which APs serve which users, to maximize the sum-rate and the $5$-percentile rate combination ($R_{\text{score}}$)
\begin{equation}
    \label{P1}
    \mathbf{P1:}\: \: \: \: \underset{A(t)}{\max}  \: \: \sum_{t=1}^T R_{\text{score}}(t), 
\end{equation}
where $A(t)$ is the joint action of all APs, \emph{i.e.}, the scheduling policy of all APs. However, optimizing the $R_{\text{score}}$ using~\eqref{r_score} in the objective directly is hard to optimize and shows instability convergence~\cite{naderializadeh2021resource}. Therefore, we adjust the objective using the proportional fairness (PF) factor~\cite{naderializadeh2021resource}. The PF factor describes the priority of each UE and is calculated as
\begin{equation}
    \label{PF_ratio}
    \text{PF}_m(t) = w_m(t) \, R_m (t),
\end{equation}
where $w_m(t)$ is a weighting factor calculated recursively as
\begin{align}
    \label{weighting_factor}
    &w_m(t) = \frac{1}{\tilde{R}_m(t)}, \\
    &\tilde{R}_m(t) = \eta R_m(t) + (1-\eta) \tilde{R}_m(t-1),
\end{align}
where $\eta$ is a step parameter and $\tilde{R}_m(0) = R_m(0)$. The PF factor is inversely proportional to the long-term rate of the user, reflecting that the higher the PF factor for a user indicates its need to be served. Therefore, as proved is~\cite{naderializadeh2021resource}, the objective in~\eqref{P1} is simplified as
\begin{equation}
    \label{P1_edited}
    \mathbf{P1:}\: \: \: \: \underset{A(t)}{\max}  \: \: \sum_{t=1}^T \sum_{m=1}^M (w_m(t))^{\lambda} \cdot R_m(t), 
\end{equation}
where $\lambda \in [0,1]$ controls the trade-off between the sum-rate and the $5$-percentile rate.

\section{Reinforcement Learning}\label{sec:RL} 
In this section, we formulate and solve the objective in~\eqref{P1_edited} using online reinforcement learning.

\subsection{Markov Decision Process}
The RRM problem can be formulated as a Markov decision process (MDP). An MDP is characterized by the tuple $\langle s_t \,a_t,r_t,s_{t+1},\gamma \rangle$, where $s_t$ is the current state, $a_t$ is the current action, $r_t$ is the reward received from taking action $a_t$ at state $s_t$ and transiting to the next state $s_{t+1}$, and $\gamma \in [0,1]$ is the discount factor that controls how much future rewards are considered in the RL problem. For a more practical and general RL formulation, we limit the number of UEs each AP can select to $K$ UEs. At the beginning of each episode, user association occurs for each AP, then, among the associated users, only the best $K$ users (the highest $K$ users in terms of the weighting factor $w_{mn}(0)$ calculated from~\eqref{weighting_factor}) are included for selection. The detailed MDP in the RRM problem is as follows
\begin{enumerate}
    \item \textbf{State:} each AP can observe two components related to each device among the selected top $K$ devices, the SINR measurement $\text{SINR}_{kn}(t)$ and the weighting factor $w_{kn}(t)$. For N available APs, the overall state of the system is 
    \begin{align}
        s_t = \big( &\text{SINR}_{11}(t), w_{11}(t), \cdots, \text{SINR}_{K1}(t), w_{K1}(t), \cdots, \\ \nonumber
        &\text{SINR}_{1N}(t), w_{1N}(t), \cdots, \text{SINR}_{KN}(t), w_{KN}(t) \big).
    \end{align}
    The state space size is $2 N K$.

    \item \textbf{Action:} each AP schedules a resource to one device only among the top $K$ devices at each time slot (or chooses to be silent and serve no UEs). The overall action space is the scheduled devices chosen by each AP, and its size is $(K+1)^N$.

    \item \textbf{Reward:} since the objective in~\eqref{P1} is hard to optimize, the reward function is formulated using the objective in~\eqref{P1_edited}
    \begin{equation}
        \label{reward}
        r_t = \sum_{m=1}^M (w_m(t))^{\lambda} \, R_m(t),
    \end{equation}
    whereas the performance is evaluated using the $R_{\text{score}}$ in~\eqref{r_score}.
\end{enumerate}

\subsection{Online RL}
RL frameworks aim to find the optimum policy $\pi^*$ that maximizes the accumulative rewards. Recently, deep neural networks provide power RL algorithms, such as deep Q-network (DQN)~\cite{mnih2015human}, proximal policy optimization (PPO)~\cite{schulman2017proximal}, and soft actor-critic (SAC)~\cite{haarnoja2018soft}, that can solve large dimension problems. We choose DQN as our online RL algorithm in this work due to its simplicity and stability~\cite{eldeeb2022multi}. In addition, it is straightforward to introduce our algorithm in the next section on top of a DQN algorithm. DQN is a model-free online RL algorithm that uses a neural network to estimate the Q-function. In addition, it is an off-policy algorithm, where previous experiences saved (in a buffer called replay memory) from previous policies are sampled to update the current Q-function.

To find the optimum Q-function, DQN updates the Bellman optimality equation by minimizing the loss
\begin{align}
\label{bellman_error}
    \mathcal{L}_{\text{DQN}} = \:& \hat{\mathbb{E}} \left[ \left(r +\gamma \max_{a^{\prime}} \hat{Q}^{(g)}(s^{\prime},a^{\prime}) 
   - Q(s,a) \right)^2 \right],
\end{align}
where $\hat{\mathbb{E}}$ is the average taken over the sampled experiences from the replay memory, $\gamma$ is the discount factor, $g$ is the gradient step (iteration), $s^{\prime}$ is the next state and $a^{\prime}$ is the action to be chosen at $s^{\prime}$. The Q-function $Q(s, a)$ is modeled using a neural network with parameters $\phi$. The main challenge in online RL is the need for continuous online interaction with the environment, which might not be feasible or safe. Next, we present offline and distributional RL as practical alternatives.

\section{Offline and Distributional RL}\label{sec:ODRL}
This section presents the proposed offline and distributional RL algorithm to solve the objective in~\eqref{P1_edited} offline.

\subsection{Offline RL}

Offline RL resorts to a static offline dataset without any online interaction with the environment. The dataset is collected using a behavioral policy, an online learning agent, a baseline policy, or even a random one. Note that, for the RRM problem, it has been proved in~\cite{10529190} that the dataset's quality heavily affects the convergence of offline RL algorithms. Deploying traditional deep RL algorithms offline fails to converge due to the distributional shift between the actions seen in the dataset and the learned actions~\cite{levine2020offline}. Conservative Q-learning (CQL)~\cite{kumar2020conservative} is a well-known offline RL algorithm that adds a regularization term (conservative parameter) to the Bellman update, overcoming the overestimation problem from the out-of-distribution (OOD) actions learned compared to those in the dataset.

Building the CQL algorithm on top of DQN architecture is straightforward, where the CQL loss is calculated as
\begin{align}
 \label{Ind_CQL_Log}
\mathcal{L}_{\text{CQL}} = &\: \frac{1}{2} \mathcal{L}_{\text{DQN}}+\alpha \hat{\mathbb{E}} \bigg[ \log \sum_{\tilde{a}}
\exp \bigl( Q(s,\tilde{a}) \bigr) 
    - \ Q(s,a)  \bigg],
\end{align}
where $\alpha > 0$ is a constant and the summation $\sum_{\tilde{a}}$ is taken over all the actions.

\subsection{Distributional RL}

Distributional RL is a variant of RL that uses the distribution over return instead of the average return while optimizing the optimum policy~\cite{bdr2023}. Quantile-regression DQN (QR-DQN)~\cite{dabney2017distributional} is a distributional RL algorithm that estimates the return distributions $Z(s, a)$ using $I$ fixed dirac delta functions. In QR-DQN, the output layer of the neural network has a size of the number of actions times the number of quantiles $I$. The distributional Bellman loss is calculated as
\begin{equation}
\label{QR_Loss}
   \mathcal{L}_{\text{QR-DQN}} = \frac{1}{I^2} \sum_{j=1}^I\sum_{j'=1}^I\zeta_{\tau_j}\left(r + \gamma \: {\theta}_{j^{\prime}}^{}(s^{\prime},a^{\prime }) - \theta_j(s,a) \right),
\end{equation}
where $\theta_j (s, a)$ is an estimate of the quantile inverse PDF of the return distribution and $\zeta_{{\tau}}$ is the quantile regression Huber loss~\cite{dabney2017distributional}.

\begin{figure}[t!]
    \centering    \includegraphics[width=1\columnwidth,trim={0cm 0 0cm 0},clip]{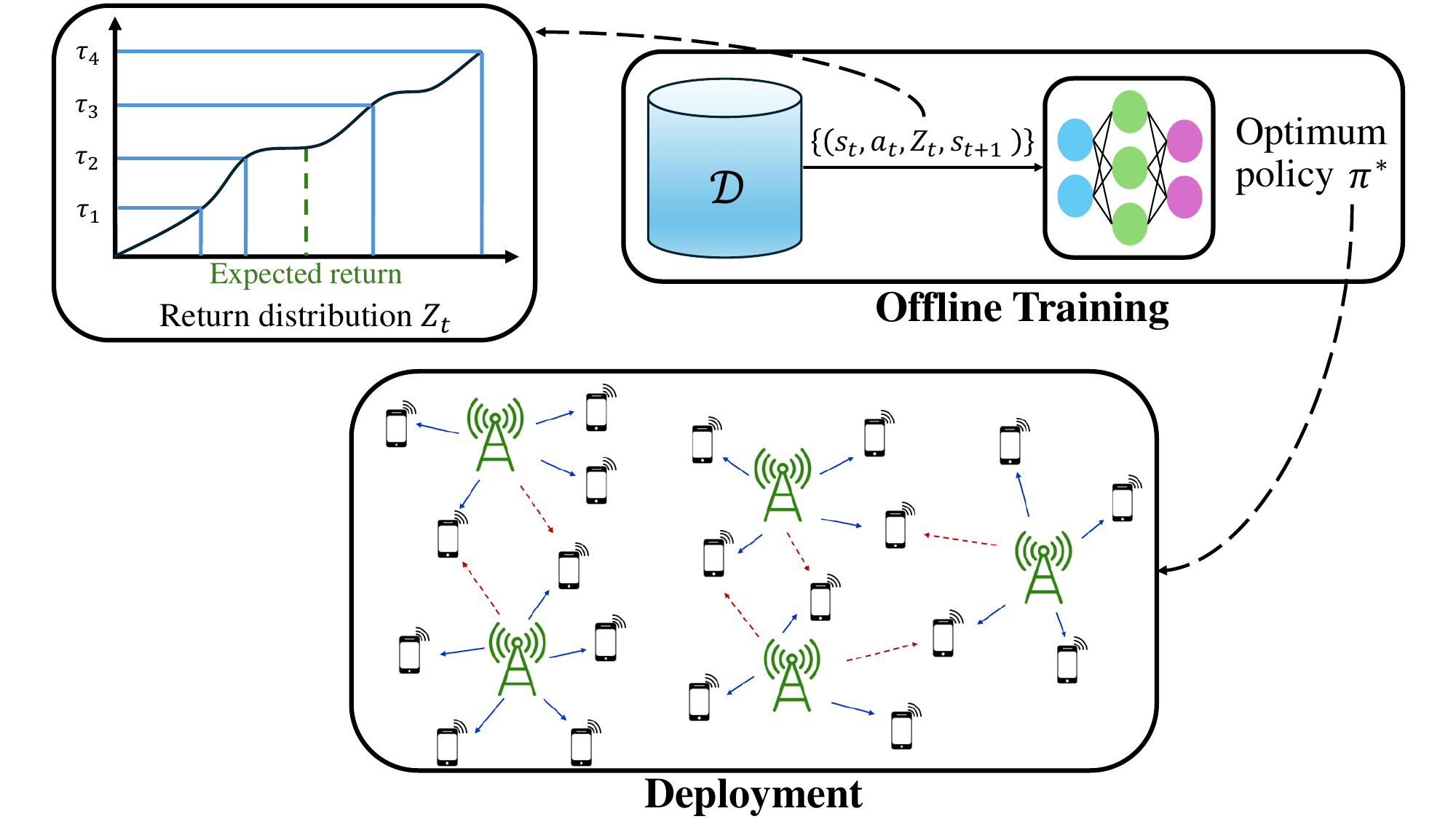} \vspace{-1mm}
    \caption{An illustrative figure for the proposed CQR algorithm.}
    \vspace{-1mm}
    \label{CQR_Fig}
\end{figure}

\subsection{Conservative Quantile Regression}
Conservative quantile regression (CQR)~\cite{eldeeb2024conser} is a variant of RL algorithms that combines CQL with QR-DQN, where the optimum Q-function is optimized offline using distributional RL. The CQR loss function is formulated as follows
\begin{align}
\label{CQR_Loss}
 \mathcal{L}_{\text{CQR}} = &\frac{1}{2} \mathcal{L}_{\text{QR-DQN}} \\ &+\alpha  \hat{\mathbb{E}}\Biggl[ \frac{1}{I} \sum_{j=1}^I \Biggl[\log \sum_{\tilde{a}} 
\exp \bigl( \theta_j(s,\tilde{a}) \bigr)
    -  \theta_j(s,a)  \Biggr] \Biggl]. \nonumber
\end{align}
Fig.~\ref{CQR_Fig} presents an illustration for the proposed CQR algorithm, and \textbf{Algorithm~\ref{CQR_Alg}} summarizes the proposed centralized training using the CQR algorithm for the RRM problem.

\begin{algorithm}[!t]
% \SetAlgoLined

    Define number of APs $N$, number of UEs $M$, number of best weighting factor users $K$, discount factor $\gamma$, learning rate $\zeta$, number of quantiles $I$, conservative penalty constant $\alpha$, number of training epochs $E$, number of gradient steps $G$, offline dataset $\mathcal{D}$, input layer size $2NK$, and output layer size $(K+1)^N I$

    %\textbf{Output:} Optimized quantile estimate $\{ {\hat{\theta}}_j(s,a)\}_{j=1}^I$

%Define $\tau_i =  \xi i/N, \ i=1,...,N$\\

Initialize network parameters

\For{\text{epoch} $e$ in $\{1$,...,$E$\}}{

\For{\text{gradient step} $g$ in $\{1$,...,$G$\}}{

Sample a batch $\mathcal{B}$ from the dataset $\mathcal{D}$

Estimate the CQR loss $\mathcal{L}_{\text{CQR}}$ in~\eqref{CQR_Loss}

Perform a stochastic gradient step based on the estimated loss with a learning rate $\zeta$

}
}

\textbf{Return} $\{\hat{\theta}_j(s,a)\}_{j=1}^I$ 

\caption{Conservative quantile regression algorithm for the RRM problem.
}
\label{CQR_Alg}  \vspace{0mm}
\end{algorithm}
 \vspace{-1mm}

\section{Experimental Results}\label{sec:results}
This section presents the numerical results of the proposed offline and distributional RL algorithm compared to the baseline models. First, we show the online RL algorithm's (DQN) performance. Then, we present the proposed offline and distributional RL algorithm.

\subsection{Baseline Schemes}
We compare the proposed algorithm to some of the state-of-the-art baseline methods:
\begin{itemize}
    \item \textbf{Random-walk:} each AP randomly chooses one of the top $K$ UEs to serve at each time slot.

    \item \textbf{Full-reuse:} each AP chooses the user with the highest PF ratio among the top $K$ UEs at each time slot.

    \item \textbf{Time-division multiplexing:} each AP serves the top $K$ UEs in a round-robin fashion. This scheme prioritizes fairness among users.
    
    \item \textbf{Information-theoretic link scheduling (ITLinQ):} each AP prioritizes its associated UEs according to their PF ratios. Afterward, each AP goes through an interference tolerance check for each UE to make sure the interference level is lower than a certain threshold $M \text{SNR}^{\eta}_{mn}$. If no UEs associated with the AP passes the interference tolerance check, this AP is turned off. This method proved to reach sub-optimal performance in the literature~\cite{6875098}.
\end{itemize}

\subsection{Simulation Parameters and Dataset}
We consider an environment of size $500$ m $\times$ $500$ m, $N = 4$ APs and $M = 24$ UEs. For the online RL, we build a DQN, where the neural network has $2$ hidden layers with $256$ neurons each. Each episode consists of $T = 2000$ time steps. We collect the offline dataset using a behavioral policy from an online DQN agent. In other words, we use the last $20 \%$ of the transitions of training an online DQN agent. We simulate a single NVIDIA Tesla V100 GPU using Pytorch framework. All the simulation parameters are shown in Table~\ref{RRM_Parameters}. To ensure scalability, we validate the proposed algorithm compared to the baselines on $100$ unique test episodes.

\subsection{Simulation Results}

\begin{figure}[t!]
    \centering    \includegraphics[width=0.95\columnwidth,trim={0cm 0 0cm 0},clip]{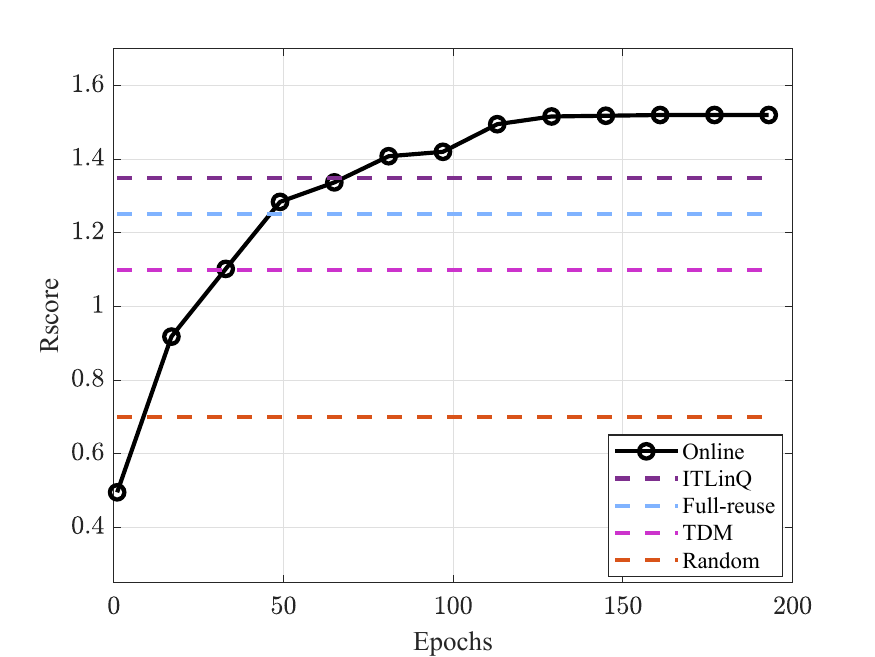} \vspace{-1mm}
    \caption{The convergence of online RL as a function of training episodes compared to the baseline methods. All the results shown are average over $100$ unique test episodes.}
    \vspace{-1mm}
    \label{Rscore_online}
\end{figure}

\begin{table}[t!]
\centering
\caption{Simulation parameters}
\label{RRM_Parameters}
\begin{tabular}{cc|cc}
\toprule
\textbf{Parameter} & \textbf{Value} & \textbf{Parameter} & \textbf{Value} \\ \midrule

$N$ & $4$ & $M$ & $24$ \\
$K$ & $3$ & $L$ & $50$ \\
$d_0$ & $10$ m & $d_1$ & $1$ m \\
$v(t)$ & $1$ m/s & $PL_o$ & $10$ dB \\
$p_t$ & $10$ dBm & $T$ & $2000$ \\
$\mu_1$ & $\frac{1}{M}$ & $\mu_2$ & $3$ \\
$\lambda$ & $0.8$ & $\gamma$ & $0.99$ \\
$I$ & $8$ & $\alpha$ & $1$ \\
$\zeta$ & $10^{-5}$ & Replay memory & $10^5$ \\
Layers & $2$ & Neurons & $256$ \\
Optimizer & Adam & Activation & ReLu \\
\bottomrule
\end{tabular} 
\end{table}

Fig.~\ref{Rscore_online} demonstrates the convergence of online RL (DQN) as a function of training episodes. We first observe that the random scheme has the worst Rscore, while TDM and full-reuse show close Rscore values. The sub-optimal scheme ITLinQ has the highest Rscore among all the baseline methods. Moreover, DQN reaches convergence (Rscore = $1.52$) after around $120$ episodes. It outperforms all the baseline schemes, including the sub-optimal scheme ITLinQ, by $20 \%$.

\begin{figure}[t!]
    \centering    \includegraphics[width=0.95\columnwidth,trim={0cm 0 0cm 0},clip]{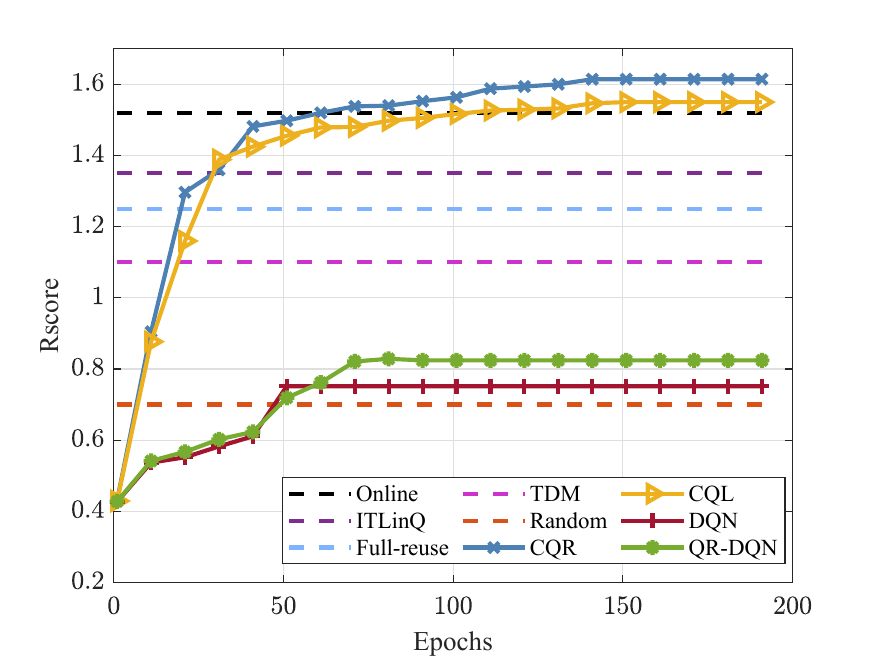} \vspace{-1mm}
    \caption{The convergence of the proposed CQR algorithm as a function of training epochs compared to other offline RL schemes and the baseline methods; the Online method is shown after convergence. All the results shown are average over $100$ unique test episodes.}
    \vspace{-1mm}
    \label{RRM_Fig}
\end{figure}

In Fig.~\ref{RRM_Fig}, we report the convergence of the proposed offline and distributional CQR algorithm compared to multiple offline/distributional RL algorithms, namely, CQL, DQN (in an offline manner), and QR-DQN. This figure also illustrates the online scheme (after convergence) and the baseline methods. Although DQN and QR-DQN achieve higher Rscore than the random scheme, they fail to converge. In contrast, CQL surpasses the conventional schemes and the online RL after convergence. However, it fails to reach the convergence of the proposed CQR algorithm, which outperforms the online RL by $20 \%$ in terms of the Rscore. This highlights the benefits of optimizing the distributional return over the average return.

\begin{figure*}[t]
    \centering
    \subfloat[ \label{sum_20}]{\includegraphics[width=0.30\textwidth]{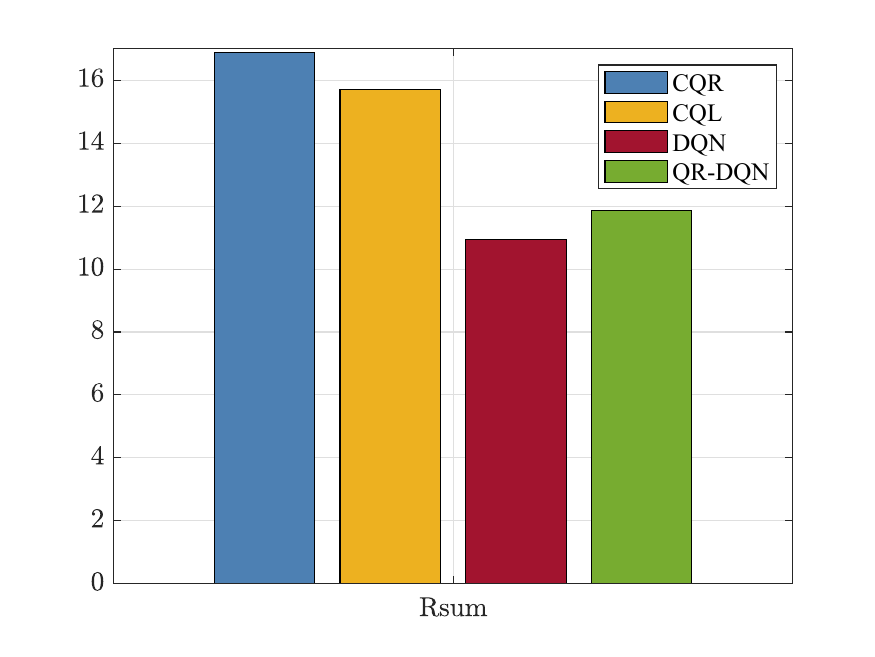}}
    \hskip -1.9ex
    \subfloat[\label{perc_20}]{\includegraphics[width=0.30\textwidth]{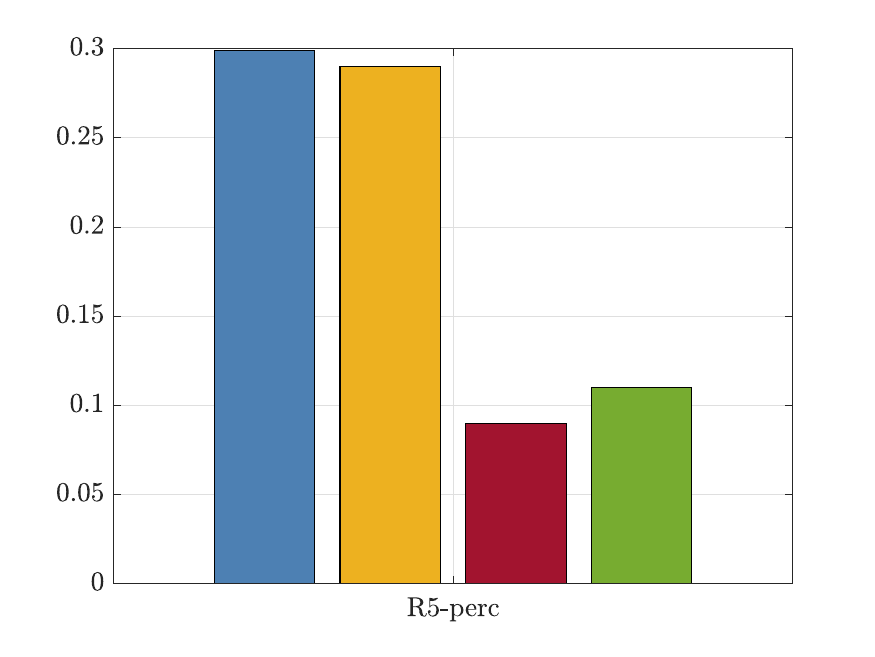}}
    \hskip -1.9ex
    \subfloat[\label{score_20}]{\includegraphics[width=0.30\textwidth]{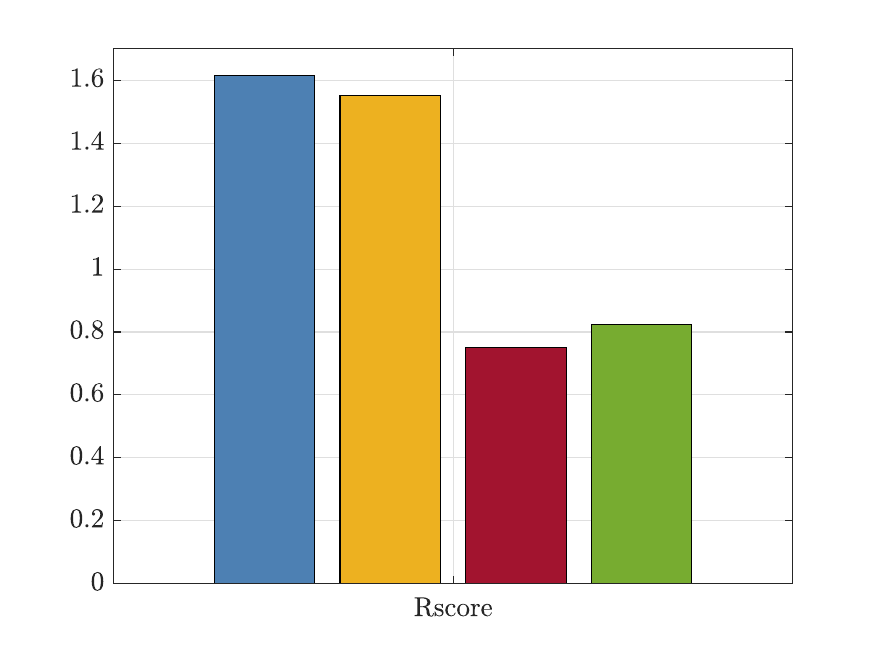}}
    \hskip -1.9ex
    \subfloat[\label{sum_10}]{\includegraphics[width=0.30\textwidth]{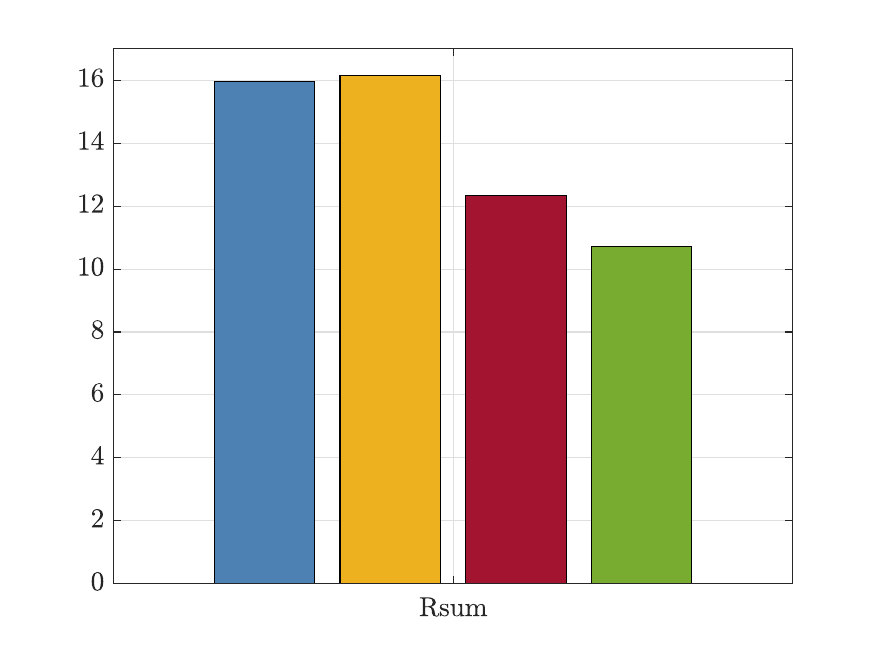}}
    \hskip -1.9ex
    \subfloat[\label{perc_10}]{\includegraphics[width=0.30\textwidth]{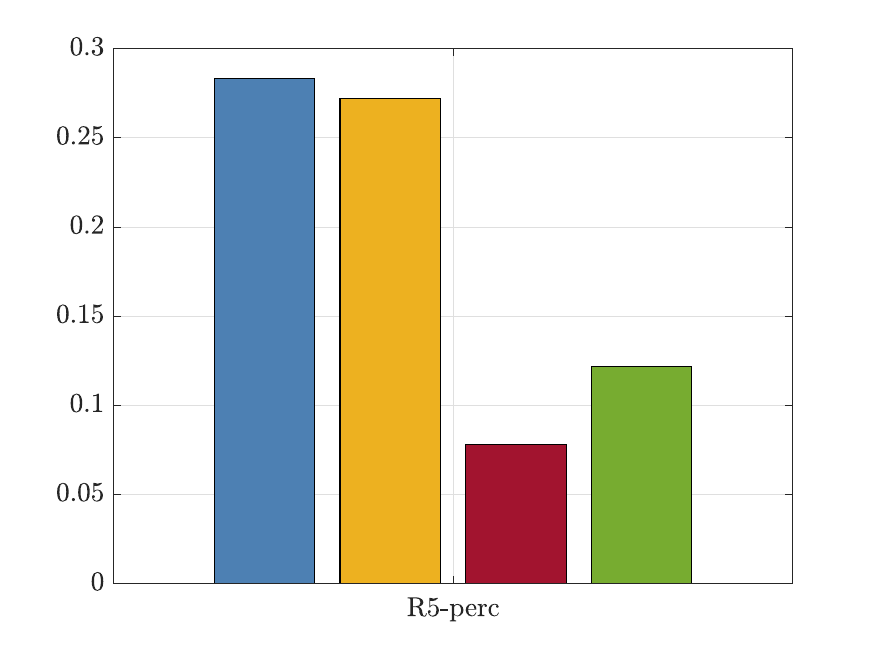}}
    \hskip -1.9ex
    \subfloat[\label{score_10}]{\includegraphics[width=0.30\textwidth]{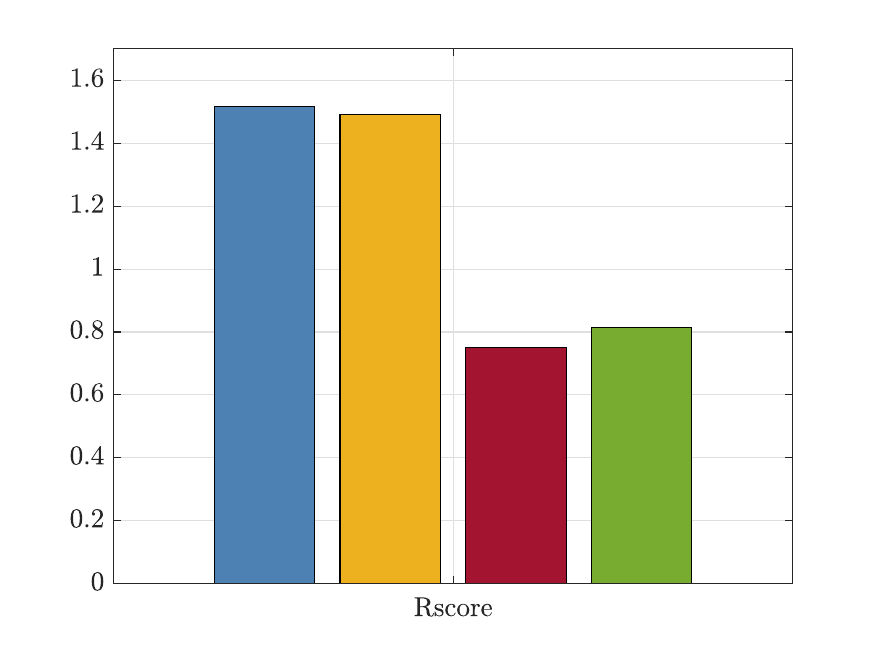}}
    \hskip -1.9ex
    \caption{The sum rate, $5$-percentile rate, and Rscore reported for the proposed CQR algorithm compared to other offline RL schemes: (a) to (c) using a dataset of $20 \%$ of the experience of online DQN and (d) to (f) using a dataset of $10 \%$ of the experience of online DQN.}
    \label{Rate} \vspace{-0mm}
\end{figure*}

Fig.~\ref{Rate} shows the performance of the proposed CQR algorithm compared to other baseline offline RL methods during testing after full training. It reports the sum-rate, $5$-percentile rate, and Rscore using an offline dataset contains $20 \%$ of the experience of a converged online DQN (Fig.\ref{sum_20}-\ref{score_20}) and using an offline dataset contains $10 \%$ of the experience of a converged online DQN (Fig.\ref{sum_10}-\ref{score_10}), respectively. Regardless of the size of the offline dataset, the CQR algorithm outperforms other offline RL schemes. In addition, the size of the dataset slightly affects the performance of CQR as the rates are higher with a larger dataset, which is often better in quality because it comes from the last experiences (good experiences) seen by a DQN agent. In the case of using a smaller dataset in training, the proposed CQR algorithm slightly sacrifices the sum-rate to increase the $5$-percentile rate, which achieves higher Rscores than the CQL algorithm, which prefers the sum-rate over the $5$-percentile rate. In addition, a significant gap is recorded between the proposed CQR algorithm and other offline schemes with smaller datasets. This highlights that the proposed model requires less data to achieve reasonable rates than other offline RL methods.

%\textcolor{purple}{HA: would it make sense to compare these to online RL? Are these figures after 120+ epochs?  }

\vspace{-1mm}
\section{Conclusions}\label{sec:conclusions} 
\vspace{-1mm}

In this paper, we developed a novel offline and distributional RL algorithm for RRM. First, we formulated the problem as MDP and then introduced the practical limitations of online RL. Afterward, we introduced the proposed model using a combination of CQL and QR-DQN. Simulation results show that the proposed model achieved a higher Rscore than all the baseline schemes. In addition, it is the only scheme to surpass online RL with a $10 \%$ gain in terms of the Rscore. Investigating the RRM problem using offline and distributional multi-agent RL is left for future work. Additionally, the proposed algorithm could be adopted for other 6G applications, such as secure intelligent reflecting surface (IRS) communication and beamformers optimization in mmWave communications.

\vspace{-1mm}
\section*{Acknowledgments} \vspace{1mm}
This work was supported by 6G Flagship (Grant Number 369116) funded by the Research Council of Finland.

\vspace{-1mm}

\bibliographystyle{IEEEtran}
\bibliography{IEEEabrv,references}

\begin{thebibliography}{10}
\providecommand{\url}[1]{#1}
\csname url@samestyle\endcsname
\providecommand{\newblock}{\relax}
\providecommand{\bibinfo}[2]{#2}
\providecommand{\BIBentrySTDinterwordspacing}{\spaceskip=0pt\relax}
\providecommand{\BIBentryALTinterwordstretchfactor}{4}
\providecommand{\BIBentryALTinterwordspacing}{\spaceskip=\fontdimen2\font plus
\BIBentryALTinterwordstretchfactor\fontdimen3\font minus \fontdimen4\font\relax}
\providecommand{\BIBforeignlanguage}[2]{{%
\expandafter\ifx\csname l@#1\endcsname\relax
\typeout{** WARNING: IEEEtran.bst: No hyphenation pattern has been}%
\typeout{** loaded for the language `#1'. Using the pattern for}%
\typeout{** the default language instead.}%
\else
\language=\csname l@#1\endcsname
\fi
#2}}
\providecommand{\BIBdecl}{\relax}
\BIBdecl

\bibitem{8714026}
N.~C. Luong, D.~T. Hoang, S.~Gong, D.~Niyato, P.~Wang, Y.-C. Liang, and D.~I. Kim, ``Applications of deep reinforcement learning in communications and networking: A survey,'' \emph{IEEE Communications Surveys \& Tutorials}, vol.~21, no.~4, pp. 3133--3174, 2019.

\bibitem{10004901}
M.~Zangooei, N.~Saha, M.~Golkarifard, and R.~Boutaba, ``Reinforcement learning for radio resource management in {RAN} slicing: A survey,'' \emph{IEEE Communications Magazine}, vol.~61, no.~2, pp. 118--124, 2023.

\bibitem{8761431}
F.~Meng, P.~Chen, and L.~Wu, ``Power allocation in multi-user cellular networks with deep {Q} learning approach,'' in \emph{ICC 2019 - 2019 IEEE International Conference on Communications (ICC)}, 2019, pp. 1--6.

\bibitem{9679392}
H.-S. Lee, D.-Y. Kim, and J.-W. Lee, ``Radio and energy resource management in renewable energy-powered wireless networks with deep reinforcement learning,'' \emph{IEEE Transactions on Wireless Communications}, vol.~21, no.~7, pp. 5435--5449, 2022.

\bibitem{naderializadeh2021resource}
N.~Naderializadeh, J.~J. Sydir, M.~Simsek, and H.~Nikopour, ``Resource management in wireless networks via multi-agent deep reinforcement learning,'' \emph{IEEE Transactions on Wireless Communications}, vol.~20, no.~6, pp. 3507--3523, 2021.

\bibitem{10066838}
N.~NaderiAlizadeh, M.~Eisen, and A.~Ribeiro, ``Learning resilient radio resource management policies with graph neural networks,'' \emph{IEEE Transactions on Signal Processing}, vol.~71, pp. 995--1009, 2023.

\bibitem{8931561}
Y.~Hua, R.~Li, Z.~Zhao, X.~Chen, and H.~Zhang, ``{GAN}-powered deep distributional reinforcement learning for resource management in network slicing,'' \emph{IEEE Journal on Selected Areas in Communications}, vol.~38, no.~2, pp. 334--349, 2020.

\bibitem{levine2020offline}
S.~Levine, A.~Kumar, G.~Tucker, and J.~Fu, ``Offline reinforcement learning: Tutorial, review, and perspectives on open problems,'' \emph{arXiv preprint arXiv:2005.01643}, 2020.

\bibitem{bdr2023}
M.~G. Bellemare, W.~Dabney, and M.~Rowland, \emph{Distributional Reinforcement Learning}.\hskip 1em plus 0.5em minus 0.4em\relax MIT Press, 2023, \url{http://www.distributional-rl.org}.

\bibitem{10529190}
K.~Yang, C.~Shi, C.~Shen, J.~Yang, S.-p. Yeh, and J.~J. Sydir, ``Offline reinforcement learning for wireless network optimization with mixture datasets,'' \emph{IEEE Transactions on Wireless Communications}, pp. 1--1, 2024.

\bibitem{3gpp_indoor}
3GPP, ``Simulation assumptions and parameters for {FDD} {HeNB} {RF} requirements,'' \emph{Tech. Rep. {R4-092042}}.

\bibitem{mnih2015human}
V.~Mnih, K.~Kavukcuoglu, D.~Silver, A.~A. Rusu, J.~Veness, M.~G. Bellemare, A.~Graves, M.~Riedmiller, A.~K. Fidjeland, G.~Ostrovski \emph{et~al.}, ``Human-level control through deep reinforcement learning,'' \emph{nature}, vol. 518, no. 7540, pp. 529--533, 2015.

\bibitem{schulman2017proximal}
J.~Schulman, F.~Wolski, P.~Dhariwal, A.~Radford, and O.~Klimov, ``Proximal policy optimization algorithms,'' \emph{arXiv preprint arXiv:1707.06347}, 2017.

\bibitem{haarnoja2018soft}
T.~Haarnoja, A.~Zhou, K.~Hartikainen, G.~Tucker, S.~Ha, J.~Tan, V.~Kumar, H.~Zhu, A.~Gupta, P.~Abbeel \emph{et~al.}, ``Soft actor-critic algorithms and applications,'' \emph{arXiv preprint arXiv:1812.05905}, 2018.

\bibitem{eldeeb2022multi}
E.~Eldeeb, J.~M. de~Souza~Sant'Ana, D.~E. P{\'e}rez, M.~Shehab, N.~H. Mahmood, and H.~Alves, ``Multi-{UAV} path learning for age and power optimization in {IoT} with {UAV} battery recharge,'' \emph{IEEE Transactions on Vehicular Technology}, vol.~72, no.~4, pp. 5356--5360, 2022.

\bibitem{kumar2020conservative}
\BIBentryALTinterwordspacing
A.~Kumar, A.~Zhou, G.~Tucker, and S.~Levine, ``Conservative {Q}-learning for offline reinforcement learning,'' in \emph{Advances in Neural Information Processing Systems}, vol.~33.\hskip 1em plus 0.5em minus 0.4em\relax Curran Associates, Inc., 2020, pp. 1179--1191. [Online]. Available: \url{https://proceedings.neurips.cc/paper_files/paper/2020/file/0d2b2061826a5df3221116a5085a6052-Paper.pdf}
\BIBentrySTDinterwordspacing

\bibitem{dabney2017distributional}
W.~Dabney, M.~Rowland, M.~Bellemare, and R.~Munos, ``Distributional reinforcement learning with quantile regression,'' in \emph{Proceedings of the AAAI Conference on Artificial Intelligence}, vol.~32, no.~1, 2018.

\bibitem{eldeeb2024conser}
\BIBentryALTinterwordspacing
E.~Eldeeb, H.~Sifaou, O.~Simeone, M.~Shehab, and H.~Alves, ``Conservative and risk-aware offline multi-agent reinforcement learning for digital twins,'' 2024. [Online]. Available: \url{https://arxiv.org/abs/2402.08421}
\BIBentrySTDinterwordspacing

\bibitem{6875098}
N.~Naderializadeh and A.~S. Avestimehr, ``{ITLinQ}: A new approach for spectrum sharing in device-to-device communication systems,'' in \emph{2014 IEEE International Symposium on Information Theory}, 2014, pp. 1573--1577.

\end{thebibliography}
\end{document}